\documentclass{article}
\usepackage{spconf,amsmath,graphicx,amssymb}

\usepackage{caption}
\usepackage{subcaption}
\usepackage{xcolor}
\usepackage{cite}
\usepackage{multirow}
\usepackage{multicol}

\title{Vision Transformer Equipped with Neural Resizer \\ on Facial Expression Recognition Task}
\name{\begin{tabular}{ccc}Hyeonbin Hwang$^{1\star}$ \quad Soyeon Kim$^{1\star}\thanks{* Equal contributions.}$ \quad Wei-Jin Park$^{2}$ \quad Jiho Seo$^{2}$ \quad  Kyungtae Ko$^{2}$ \quad Hyeon Yeo$^{1\dagger}\thanks{† Corresponding author}$\end{tabular}}

\address{$^{1}$KAIST, Korea \quad \quad
$^{2}$ACRYL, Korea \quad \quad}
\begin{document}
%
\maketitle
\begin{abstract}
    When it comes to wild conditions, Facial Expression Recognition is often challenged with low-quality data and imbalanced, ambiguous labels. This field has much benefited from CNN based approaches; however, CNN models have structural limitations to see the facial regions in distance. As a remedy, Transformer has been introduced to vision fields with a global receptive field but requires adjusting input spatial size to the pretrained models to enjoy its strong inductive bias at hands. We herein raise the question of whether using the deterministic interpolation method is enough to feed low-resolution data to Transformer. In this work, we propose a novel training framework, Neural Resizer, to support Transformer by compensating information and downscaling in a data-driven manner trained with loss function balancing the noisiness and imbalance. Experiments show our Neural Resizer with F-PDLS loss function improves the performance with Transformer variants in general and nearly achieves the state-of-the-art performance.
\end{abstract}
\section{Introduction}
\label{sec:intro}

 Facial expressions recognition (FER) has recently drawn much attention and has accomplished tremendous achievements thanks to the advancement of deep learning techniques. However, when it comes to in-the-wild (ITW) scenarios, several inherent problems are posed that make this task more challenging than classic vision benchmarks \cite{li_deng_2020, li2017reliable}.

First of all, datasets obtained from crowdsourcing inevitably elicit large image variance as in size, alignment, occlusion, and pose \cite{noisylabelsurvey21}. To fully utilize a convolutional neural network(CNN), training images are often preprocessed along with normalization techniques to suppress other variations. Nevertheless, the inherent ambiguity on individual facial expression become more evident on low-quality data, smoothing out the fine-grained features which contribute significantly to the overall performance. 

The second challenge is the inherent problem of the FER task originating from the human subjectivity when annotating \cite{barsoum2016training,li2017reliable} – noisy labels and class imbalance. Compared to annotating some discrete objects in an image, facial expression classes can not be consistently annotated by individuals with different backgrounds. Meanwhile, most of the real-world datasets inevitably face imbalanced classes, resulting in a bias against the minor classes \cite{hupont2019demogpairs}. While careful data filtering strategies may help, given that the FER task generally has a limited number of data, one should take a risk of having either fewer training examples for major classes or limited augmented data for minor classes.

Several previous works have focused to resolve the aforementioned challenges with CNN based approach. In respect to the architecture, considering that the FER task heavily depends on fine-grained key features, CNN with attention network \cite{wang2019region, wang2020suppressing} was introduced to give more weights for the important regions. Based on such CNN models, \cite{ngo2020facial} devised a weighted loss on each class’s relative portion of the total dataset, \cite{wang2019co} used loss values as cues to catch the outliers and the noisy data samples, and \cite{psr20} proposed adaptive label smoothing loss to resolve different confusion distribution of labels inspired by \cite{ lin2017focal , cao2019learning}. 
 
Orthogonal to these works, recent studies explore the applicability of Vision Transformer owing to its advantages on global reception fields with self-attention \cite{aouayeb2021learning, li2021mvt,robust21}. However, one of the drawbacks is that images must be all uniformly resized, generally \textit{upscaled from the low-resolution}, to fit into the pretrained Transformer input spatial size. However, such practice may not be optimal because the recommended approach for adoption in the downstream target task is to use a higher resolution than the pretrained dataset \cite{dosovitskiy2021image}.

\begin{figure*}[t]
     \centering
     \begin{subfigure}[b]{0.25\columnwidth}
         \centering
         \includegraphics[width=\linewidth]{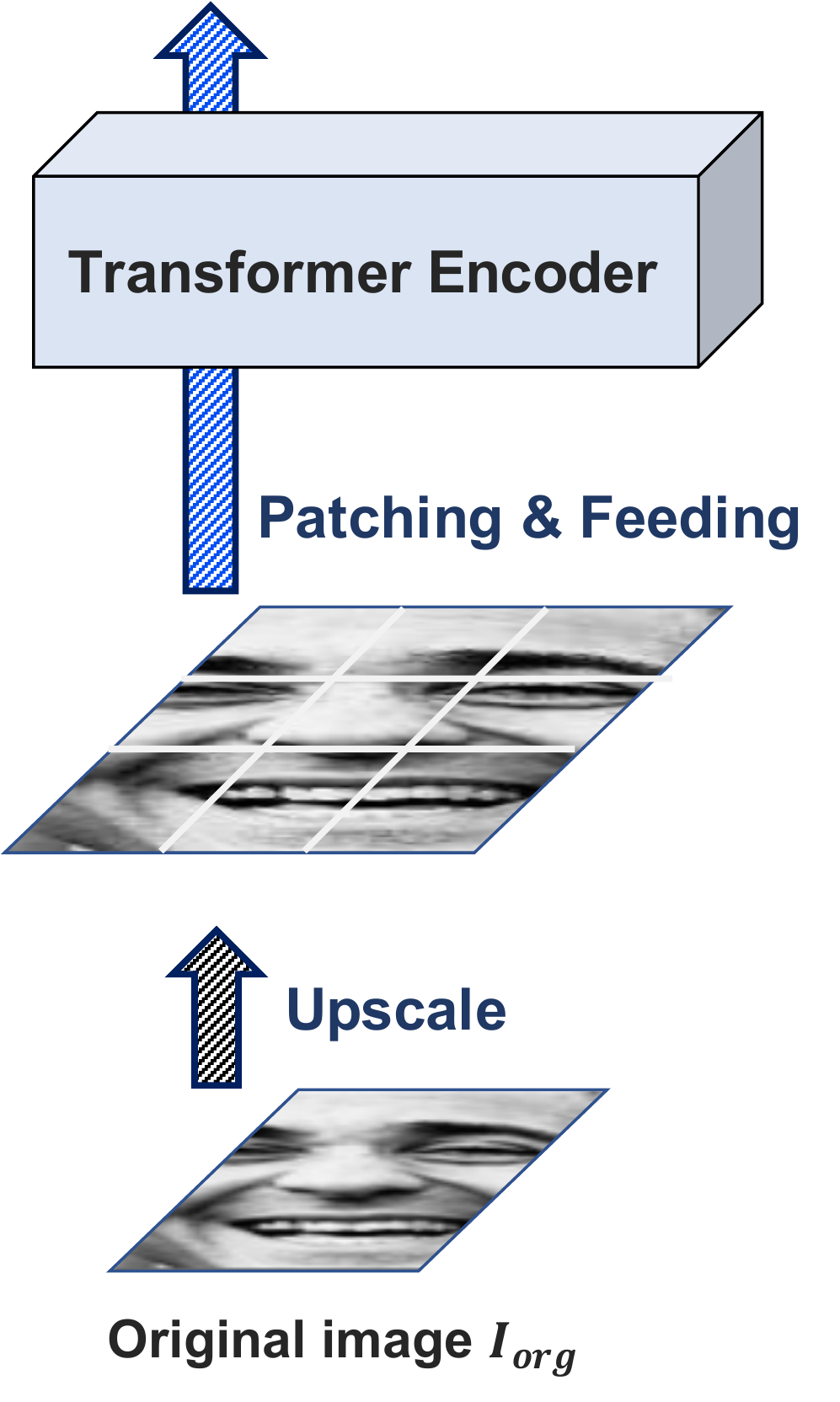}
         \caption{Baseline.}
         \label{fig:y equals x}
     \end{subfigure}
     \hfill 
     \begin{subfigure}[b]{0.59\columnwidth}
         \centering
         \includegraphics[width=\linewidth]{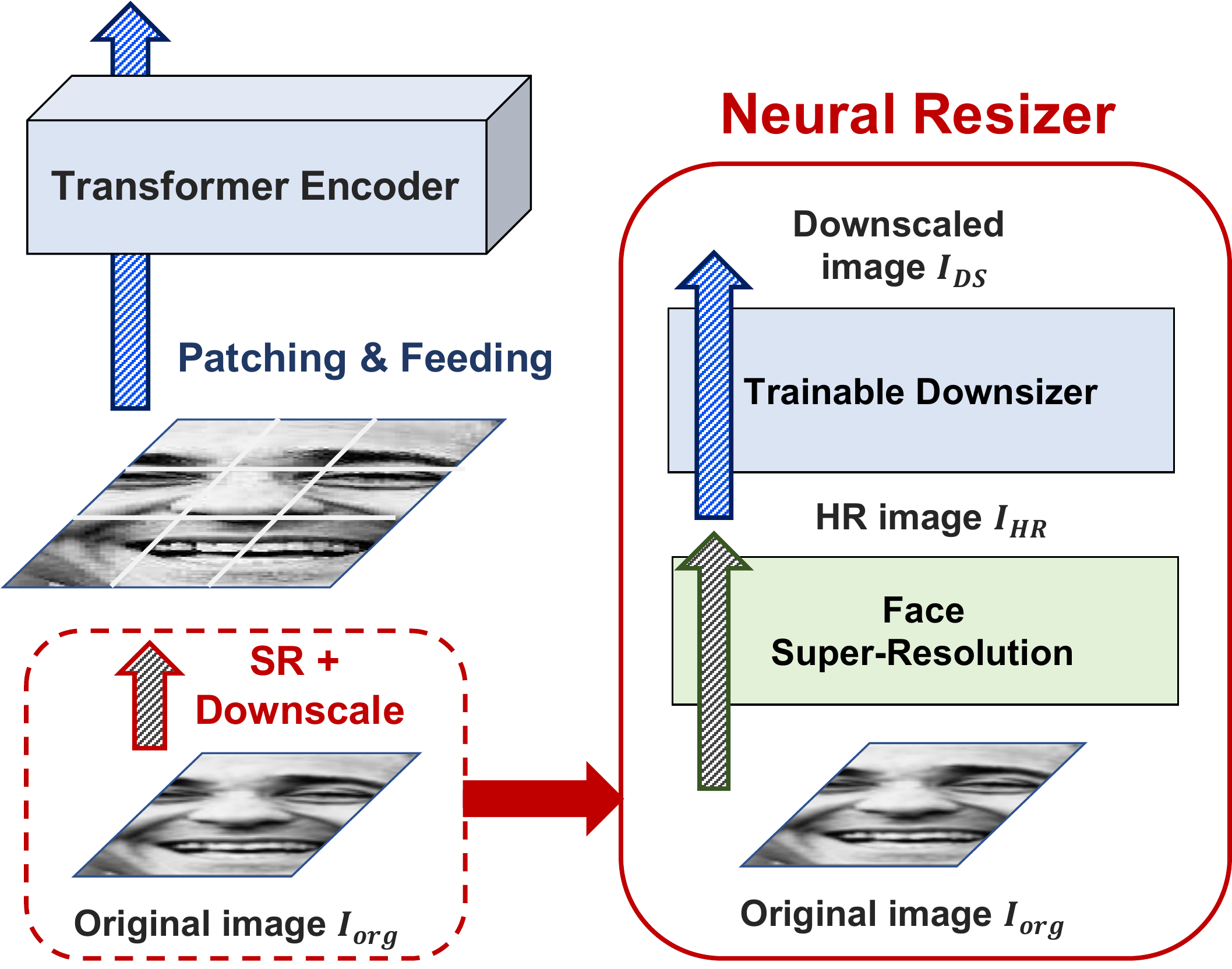}
         \caption{Proposed training pipeline.}
         \label{fig:three sin x}
     \end{subfigure}
     \hfill 
     \begin{subfigure}[b]{\columnwidth}
         \centering
         \includegraphics[width=\linewidth]{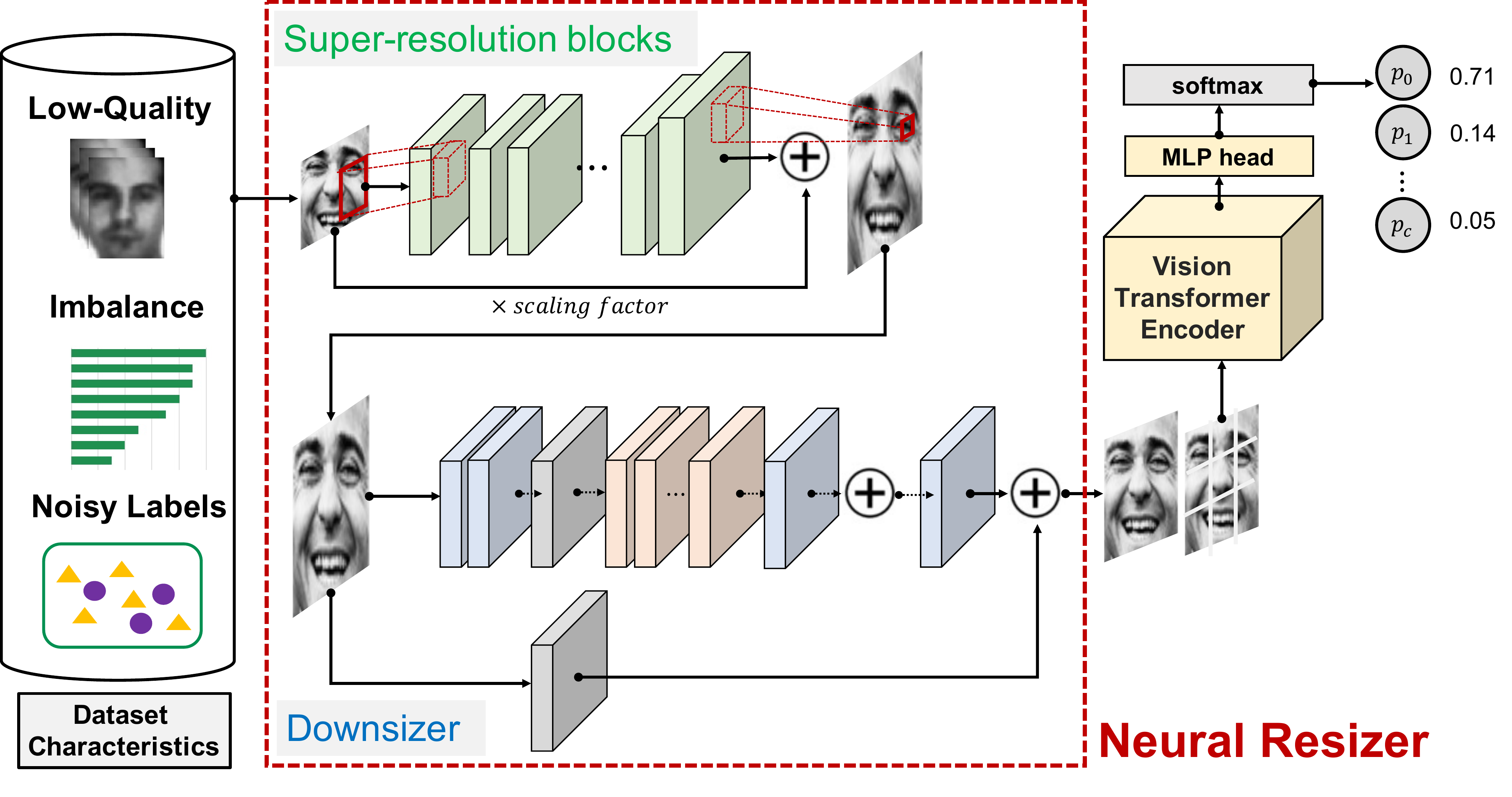}
         \caption{Details of Neural Resizer.}
         \label{fig:three sin x}
     \end{subfigure}
     
        \caption{\textbf{(a),(b)} The comparison of the naive use case and the proposed framework, Neural Resizer, to apply vision transformer variants in FER Task. \color{blue} Blue blocks : \color{black} are only involved in parameter updating. \textbf{(c)} Before feeding to Transformer directly, we organize a two-step process, which first maps low-resolution images to a high resolution followed by a trainable downsizer to match the fixed input spatial size for Transformer. }
        \label{fig:concept}
\end{figure*}

\textbf{Contribution.} Our contribution is summarized to three-fold: \textit{(i)} We propose a novel training framework to fully provide Vision Transformer variants with strong generalization power by a trainable resizing module, Neural Resizer. \textit{(ii)} We propose a novel loss function that puts importance on minor as well as ambiguous classes adaptively. \textit{(iii)} We empirically show that one of the SOTA transformer, Swin-Transformer\cite{liu2021swin}, achieves comparable performance to the state-of-the-art CNN based approaches when jointly trained with our proposed framework.  

\section{Proposed Architecture}
\subsection{Neural Resizer for Vision Transformer}

Our goal throughout Neural Resizer(NR), $f_{\Theta}$ , is to feed low-quality image in a way that Vision Transformer can well adapt to a domain-specific task. Inspired by previous works on trainable downscaling \cite{kim2018task,talebi2021learning}, we thoroughly follow the generalized formulation including the compensation process to enrich the image information. Let $I_{org}$, $I_{hr}$, $I_{ds}$ denote original image, high-resolution image and down-scaled image where each of the spatial size is $I_{org} \in R^{H \times W \times C}$, $I_{hr} \in R^{hH \times hW \times hC}$, $I_{ds} \in R^{tH \times tW \times tC}$ respectively and $h,t$ is the scaling factor for the high-resolution and target size. Then, the Neural Resizer is formulated as an approximate mapping function $f_{\Theta}: R^{H \times W \times C} \mapsto R^{tH \times tW \times tC}$. To train  $f_{\Theta}$, we decompose the function as
\begin{equation} \label{eq:NR1}
    f_{\Theta} = g_{\theta} \circ T(f_{sr}(I_{org})) \circ g_{\phi} + T(f_{sr}(I_{org}))
\end{equation} 
where $f_{sr}, T$ denote a pretrained super-resolution function and a conventional deterministic interpolation method which do not involve in  updating parameter process respectively, and $\Theta \in \{\theta, \phi \}$  denotes trainable parameters. In practice, $f_{sr}$ can be any type of off-the-shelf pretrained super resolution function and $g_{\theta}, g_{\phi}$ can be implemented using simple stacked CNN layers.

\subsection{Learning with Noisy and Imbalanced Labels}

In the real world, not only accuracy of annotation by humans but the balance between the classes is also hardly guaranteed. To minimize human error, \textit{voting} is often employed to annotate each data by multiple annotators to construct a class distribution.
\cite{psr20} predefined voting information and class distribution of datasets as prior knowledge and implemented modified label smoothing (Prior Distribution Label Smoothing, PDLS) loss function to overcome conventional Cross Entropy which applies to only a single class with the highest vote count.

Nevertheless, PDLS still overlooked the imbalance issue. To handle both of the aforementioned issues, we propose a novel loss function called F-PDLS(Focal PDLS), which applies noise distribution to the classifying task while compensating noise balance to improve accuracy additionally, inspired by \cite{lin2017focal}. Our loss is designed to utilize prior knowledge of the label's confusion and focus on difficult tasks(classes) to classify correctly so that the model can cover the imbalance of labeled ITW datasets as shown in Eq.\ref{eq1} and Eq.\ref{eq2}.

\begin{equation}\label{eq1}
L_{F-PDLS}=-\sum_{c\in C}^{} (1-\sigma(z_c))^\gamma*L_{PDLS}^{c}         \end{equation}
\begin{equation}\label{eq2}
L_{PDLS}^{c}=(t_c*\alpha+d_{kc}*(1-\alpha))*log(\sigma(z_c) )  \end{equation}
where $z_c$ is the output of the model and $L_{PDLS}^{c}$ is PDLS loss for each class $c\in C$ respectively.

All notations in Eq.\ref{eq1} are similar to PDLS loss function in \cite{psr20} except a modulating factor to the PDLS loss expressed as $(1-\sigma(z_c))^\gamma$ where $\gamma$ is a positive-valued focusing parameter.

\section{Experiments}
\subsection{Datasets}

\textbf{FERPlus}  consists of 28,709 training images, 3,589 validation images
and 3,589 test images. They are all in grayscale and size of 48 $\times$ 48. Each image was manually assigned to one of 8 predefined expression classes - neutral, happiness, surprise, sadness, anger, disgust, fear, contempt - by in total of 10 annotators. 

\textbf{RAF-DB} contains 29,672 facial images annotated with basic or compound expressions independently labeled by 40 annotators. It only consists of seven different expressions, excluding contempt. For our experiment, we use aligned RAF-DB, where all 12,271 training images and 3,068 test images are resized to 100 $\times$  100.

\begin{table}
\centering
\caption{Comparison with various state-of-the-art \textbf{small-sized} Transformers on FERPlus, tested with sole backbone architecture and our framework }
\begin{tabular}{ccc} 
\hline
\textbf{Models} & \textbf{CE + Vanilla} & \textbf{F-PDLS + Proposed}\\ 
\hline
ViT    \cite{dosovitskiy2021image} & 88.84       & 88.87                 \\
DeiT \cite{touvron2021training} &    88.00        & 88.09              \\   
ConViT \cite{dascoli2021convit}  &  88.12     & 88.53              \\
XCiT \cite{elnouby2021xcit}      &  88.22   & 88.81              \\ 
\hline
\textbf{Swin-S} & 88.69 & \textbf{89.28}     \\
\hline
\end{tabular}
\label{tab:table1}
\end{table}

\subsection{Implementation Details}
As our backbone architecture, we utilize the pretrained Transformer families released from the official library with the expected image resolution of 224. We used Deep Iterative Collaboration(DIC) model\cite{ma2020deep} pretrained with facial dataset(e.g, Helen\cite{le2012interactive}) for super-resolution, while in terms of the resizer, we follow the previous work configuration \cite{talebi2021learning}. For the hyperparameter setting, we trained our model for 100 epochs with a batch size of 64. We set the initial learning rate to 1e-4, while decreasing it one quarter every 10 epochs. We empirically employ constructive data augmentation techniques because the FER task is sensitive to certain types of augmentation such as rotation. To evaluate on FERPlus, we follow the test setting referred in \cite{barsoum2016training}, and since its images consist of 48$\times$48 spatial size, we upscale them into high-resolution of 384$\times$384. In contrast, RAF-DB dataset has 100$\times$100 size images, so we first downscale them to 48$\times$48 and after upscale to 384$\times$384 due to internal API issues.

\subsection{Evaluation}
\subsubsection{Main Results}
Table.\ref{tab:table1} shows the overall result of our proposed framework tested with various backbone Transformers. We observe a general trend in the increase of accuracy with our proposed framework, demonstrating its broad applicability with many Transformer-based variants. Also, Table.\ref{tab:table4} confirms this idea as the performance increases regardless of the loss function choice. Specifically, while ViT\cite{dosovitskiy2021image} has the best accuracy in the vanilla setting, we observe that Swin Transformer\cite{liu2021swin} equipped with Neural Resizer and F-PDLS achieves the best accuracy. We conjecture that our approach is well compatible with Swin Transformer(Swin-T) architecture which flexibly observes features map at various scales. In other words, its sequential fine-unit patch merging step can distinguish the fine-grained features more attentively by our proposed approach. Therefore, we use Swin-T in the following experiments and denote Swin-S, Swin-B, and Swin-L for Swin-Small, Swin-Base, and Swin-Large, respectively.

\begin{table} 
\centering
\caption{Ablation study on the effect of each module, when \textbf{downscaling} and \textbf{upscaling} images, tested with \textbf{Swin-S}, on FERPlus, using F-PDLS}
\begin{tabular}{cccccc} 
\hline
Setting    & model & STN & Up.     & Down.    & Acc.     \\ 
\hline
a        & Swin-S & - & Bi.    & -   &  88.69       \\
b      & Swin-S & - & Bi.    & LTR          &  88.53        \\
c      & Swin-S & - & SR          & Bi.     & 89.03 \\
d &  Swin-S & - &   SR      & LTR        & 89.28  \\ 
\hline
\textbf{e} &  \textbf{Swin-B} & \textbf{\checkmark} &   \textbf{SR}          & \textbf{LTR}          & \textbf{89.50}   \\ 
\hline
\end{tabular}
\label{tab:table2}
\end{table}

\begin{table}
\centering
\caption{Comparison across the effect of the  loss function, tested on both Vanilla Swin-T and Proposed architecture, \textbf{with Swin-B}}
\begin{tabular}{ccccc} 
\hline
Loss    & Vanilla  & Proposed        \\ 
\hline
Cross-Entropy &    88.72       &  88.87        \\
PDLS \cite{psr20}&  88.69      & 88.91  \\
\hline
\textbf{F-PDLS (ours)} &  \textbf{88.78}     & \textbf{89.50} \\
\hline
\end{tabular}
\label{tab:table4}
\end{table}

\subsubsection{Effectiveness of Each Module}

  \textbf{Quantitative Results } Table.\ref{tab:table2} shows the ablation study of replacing each module with deterministic bilinear interpolation. When using Swin-S, our method \textbf{(d)} outperforms interpolation based traditional approach \textbf{(a)} by \textbf{0.59\%} in total. It can be concluded that Learning-To-Resize Module(LTR)\cite{talebi2021learning} itself has limitations when applied to interpolated low resolution data, but works better when downsized from the higher resolution with more information. The gist from these results is the importance of data quality before neurally downsizing. Thus, we exploit a simple module, Spatial Transformer Network \cite{stn}, to align the necessary features before feeding to Super Resolution (SR) modules as an extended study. Even though the depth of Swin-T does not show a strong correlation with the maximum accuracy attained, there is a tendency towards slight improvement for deeper models. Thus, we select the middle level Swin-B when applying STN. Setting \textbf{(e)} shows our best result, which outperforms our base case \textbf{(a)} by \textbf{0.81\%}.
  
\begin{figure}
\begin{center}
\includegraphics[width=0.9\linewidth]{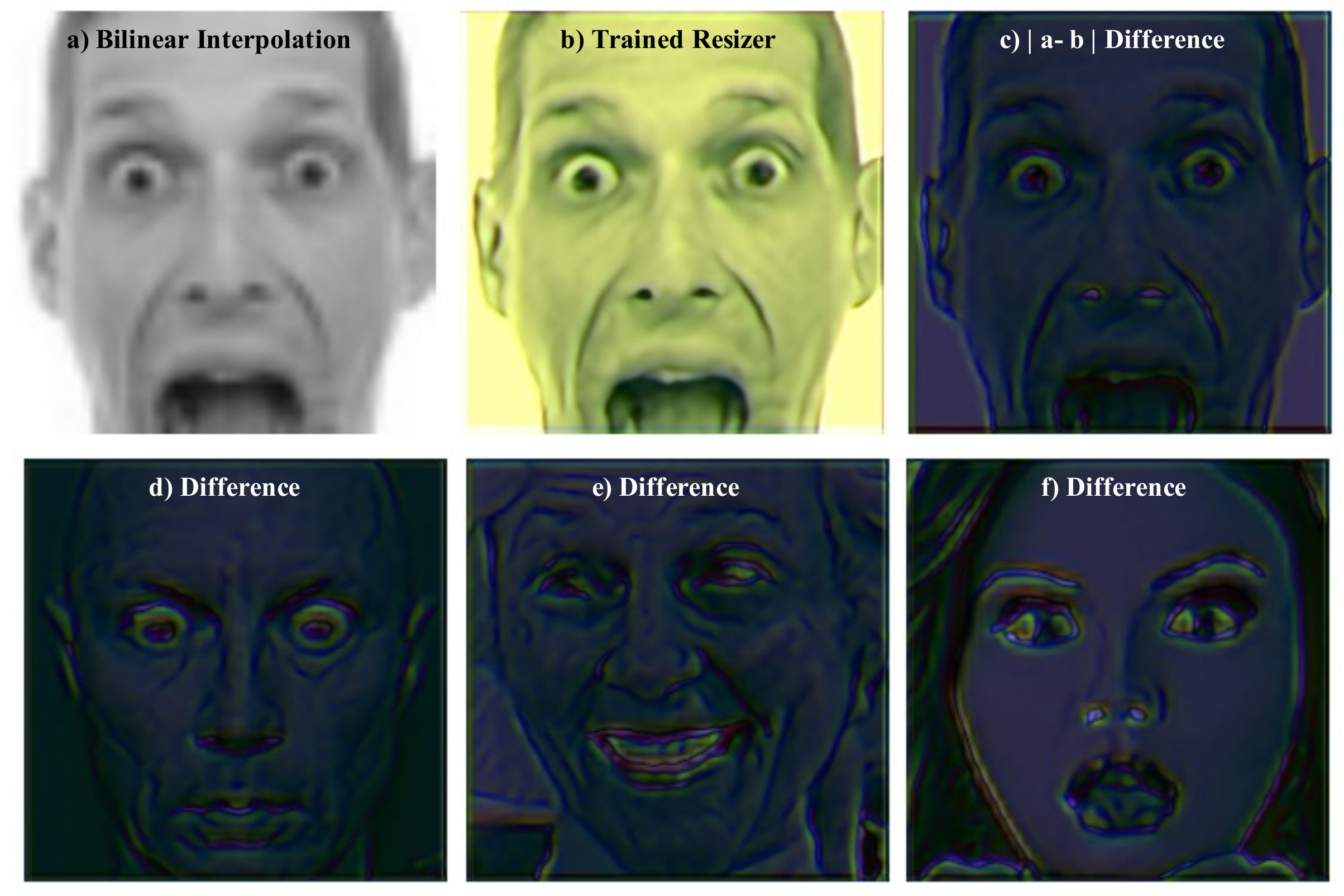}
\end{center}
   \caption{Example of the trainable resizer output. \textbf{First row :}  the result of deterministic resizer(e.g, bilinear interpolation), the proposed trained resizer and the absolute difference between (a) and (b) from the left. \textbf{Second row:} More examples of the difference.}
\label{fig:examples}
\end{figure}

\textbf{Qualitative Results } To visually deliver the role of NR, the examples in Figure \ref{fig:examples} presents that our framework successfully captures fine-grained features like the line of the wrinkles compared to the deterministic interpolation approaches. That is, image shape is not notably changed, but the edges of the discriminant features are more conspicuously accented which facilitates the classification process. Compared to the previous work\cite{robust21} which applies typical vision method(e.g, LBP\cite{shan2005robust}) and attentional selective fusion branch, we consider our trained resizer plays the incorporated role in one shot. In addition, it can save significant computing and memory footprint while training, as we only need a single branch(NR) to train and a single image to feed to NR before Transformer while the previous work has to feed both original and LBP-applied images and needs to train dual branches.

\subsubsection{The Effectiveness of Loss}
    In addition to compensating for low-quality data, our other goal is to mitigate the inherent problems of FER task - class imbalance and noisy labels. To demonstrate the effectiveness of F-PDLS to the problems, we conduct the experiments on Cross-Entropy and Prior-Distribution Label Smoothing\cite{psr20} loss function as the comparative standards under the proposed setting \textbf{(e)} in Table.\ref{tab:table2}. Interpreting the results in Table.\ref{tab:table4}, while F-PDLS achieves the highest accuracy with a trivial increase of \textbf{0.06\%}, there is no notable difference between the loss functions when used in vanilla Swin-T. However, the gain on Transformer equipped with NR increases to \textbf{0.63\%}. To conclude with the empirical results, we hypothesize that our NR plays a role as a magnifier to F-PDLS which puts importance on minor and ambiguous samples. In other words, NR helps the Transformer to look closely at the fine-grained features which were ignored when the only image itself is fed to the Transformer directly.

\subsubsection{Comparison With State-Of-The-Art Networks in FER}
Table.\ref{tab:table5} shows our performance in two ITW datasets compared to other state-of-the-art approaches in the FER task including the recent transformer-based networks. To the best of our knowledge, our model outperforms all of the transformer-variants and achieves the second highest performance on FERPlus with \textbf{89.50\%}, and \textbf{88.57\%} on RAF-DB.

\section{Conclusion}
In this paper, we propose a novel training framework to leverage Transformer under the realistic FER task condition with the proposed trainable resizer and a loss function called Neural Resizer and F-PDLS respectively. Neural Resizer first restores the image to a higher resolution to leverage enhanced information in low-quality image data. Then, the enriched image is downscaled in a data-driven manner by the downsizer. By jointly optimizing the downsizer with Transformer with F-PDLS, which puts importance on minor as well as ambiguous classes, the proposed framework helps Transformer to adapt to the target task better. The experiments suggest that our framework, as well as loss function, have improved the performance of Transformer variants in general. Furthermore, we empirically show that Swin-Transformer achieves the competitive results compared to the strong baselines.

\begin{table}
\centering
\caption{Comparison with other state-of-the-art methods for In-the-wild FER task. \textbf{*} denotes accuracy trained with Swin-Large }
\begin{tabular}{cccc} 
\hline
Type                         & Method               & FERPlus              & RAF-DB                \\ 
\hline
\multirow{3}{*}{\textbf{CNN}} & RAN \cite{wang2019region}           &       89.16           &   86.90       \\
                             & SCN \cite{wang2020suppressing}         &       89.35          &    88.14     \\               
                             & PSR \cite{psr20}        &        89.75   &         88.98
                             \\ 
\hline
\multirow{4}{*}{\textbf{Transformer}} & LBP + CVT \cite{robust21}            & 88.81                & 88.14                 \\
                             & MVT  \cite{li2021mvt}               & 88.88                & 87.03                 \\
                             & VIT + SE \cite{aouayeb2021learning}            & -                    & 86.18                 \\
                             & \textbf{ours}        & \textbf{89.50}       &    \textbf{88.57}${^\textbf{*} }$              \\ 
\hline
\end{tabular}
\label{tab:table5}
\end{table}
\bibliographystyle{IEEEbib}
\bibliography{main}

\end{document}